\documentclass[runningheads]{llncs}
\usepackage{makeidx}
\usepackage{graphicx}
\usepackage{amsmath,amssymb} %
\usepackage{color}
\usepackage{grffile} %
\usepackage{multirow} %
\usepackage[caption=false]{subfig}
\usepackage[lined]{algorithm2e} %
\usepackage{url}

\graphicspath{{figs/}}

\DeclareMathOperator{\dimfunc}{dim}

\begin{document}
\pagestyle{headings}
\mainmatter

\def\GCPR16SubNumber{80}

\title{Depth Estimation Through a\\ Generative Model of Light Field Synthesis}

\titlerunning{Depth Estimation Through a Generative Model of Light Field Synthesis}
\authorrunning{Mehdi S. M. Sajjadi \and Rolf K{\"o}hler \and Bernhard Sch{\"o}lkopf \and Michael Hirsch}
\author{Mehdi S. M. Sajjadi \and Rolf K{\"o}hler \and Bernhard Sch{\"o}lkopf \and Michael Hirsch \\
        \{msajjadi, rkoehler, bs, mhirsch\}@tuebingen.mpg.de }
\institute{Max-Planck-Institute for Intelligent Systems\\
           Spemannstra{\ss}e 38, 72076 T{\"u}bingen, Germany}
\maketitle
\vspace{-6mm}
\begin{abstract}
  Light field photography captures rich structural information
  that may facilitate a number of traditional image processing
  and computer vision tasks. A crucial ingredient in such
  endeavors is accurate depth recovery.	 We present a novel
  framework that allows the recovery of a high quality
  continuous depth map from light field data. To this end we
  propose a generative model of a light field that is fully
  parametrized by its corresponding depth map.	The model
  allows for the integration of powerful regularization
  techniques such as a non-local means prior, facilitating
  accurate depth map estimation. %
  Comparisons with previous methods show that we are able to
  recover faithful depth maps with much finer details.	In a
  number of challenging real-world examples we demonstrate
  both the effectiveness and robustness of our approach.
\vspace{-5mm}
\end{abstract}
\section{Introduction}
\vspace{-3mm}
\makeatletter{\renewcommand*{\@makefnmark}{}
\footnotetext{The final publication is available at Springer via \url{http://dx.doi.org/10.1007/978-3-319-45886-1_35}.}\makeatother}
Research on light fields has increasingly gained popularity driven
by technological developments and especially by the launch of the Lytro
consumer light field camera \cite{ng2006digital} in 2012.
In 2014, Lytro launched the \emph{Illum} follow-up model.
While these cameras are targeted to the end-consumer market, the
company Raytrix \cite{perwassnext} manufactures high-end light field
cameras, some of which are also capable of video recording, but aimed
for the industrial sector. Both Lytro and Raytrix use an array of
microlenses to capture a light field with a single camera.

Prior to the first commercially available light field cameras, other
practical methods have been proposed to capture light fields, such as a
camera array \cite{vaish2004using}, a gantry robot mounted with a
DSLR camera that was used to produce the Stanford Light Field dataset
\cite{stanfordLFdataset} or a programmable aperture which can be used
in conjunction with a normal 2D camera to simulate a light field camera
\cite{liang2008programmable}.

Previously proposed approaches
\cite{kim2013scene,tao2013depth,diebold2013epipolar}
model a depth map as a Markov Random Field and cast depth estimation
as a multi-labelling problem, so the reconstructed depth map consists of
discrete values. In order to keep the computational cost to a
manageable size, the number of depth labels is typically kept low which
results in cartoon-like staircase depth maps.

In contrast, we propose an algorithm which is capable of
producing a continuous depth map from a recorded light field and which hence
provides more accurate depth labels, especially for fine structures.
Before we discuss related work in
Section~\ref{sec:related} and present our method in
Section~\ref{sec:overview}, we lay down our notation and light field
parametrization in the following section.
\section{Light field parametrization} 
\vspace{-3mm}

It is instructive to think of a light field as a collection of images
of the same scene, taken by several cameras at different
positions. The 4D light field is commonly parametrized using the two
plane parametrization introduced in the seminal paper by
\cite{levoy1996light}. It is a mapping
\begin{equation*}
  L : \Pi \times \Omega \to \mathbb{R} \qquad \qquad (s,t,x,y) \mapsto L(s,t,x,y)
\end{equation*}
where $\Omega \subset \mathbb{R}^2$ denotes the image plane and $\Pi \subset
\mathbb{R}^2$ denotes the focal plane containing the focal points
of the different virtual cameras. $(x,y)$ and $(s,t)$ denote points in the image plane
and camera plane respectively. In a discretely sampled
light field, $(x,y)$ can be regarded as a pixel in an image and
$(s,t)$ can be regarded as the position of the camera in the grid of
cameras. We store the light field as a 4D object with $\dimfunc(L) =
(S,T,X,Y)$.

For the discrete case, also called \emph{light field photography},
each virtual camera is placed on a cross-section of an equispaced $n
\times n$ grid providing a view on the same scene from a slightly
different perspective.
Two descriptive visualizations of the light field arise if different
parameters are fixed. If $s=s^*$ and $t=t^*$ are fixed, so-called
\emph{sub-aperture images} $I_{s^*,t^*}$ arise. A
sub-aperture image is the image of one camera looking at the
scene from a fixed viewpoint $(s^*,t^*)$ and looks like a normal 2D
image. If ($y=y^*$, $t=t^*$) or ($x=x^*$, $s=s^*$) are kept fixed,
so-called \emph{epipolar plane images} (EPI) \cite{bolles1987epipolar} $E_{y^*,t^*}$ or $E_{x^*,s^*}$
arise. %

An interesting property of an EPI is that a point $P$ that is
visible in all sub-aperture images is mapped to a straight line in the EPI, see
Fig.~\ref{fig:overview_refinement_epi}. This characteristic
property has been employed for a number of tasks such as denoising
\cite{goldluecke2013variational,dansereau2013light}, in-painting
\cite{goldluecke2013variational}, segmentation
\cite{wanner2013globally,li2014saliency}, matting
\cite{cho2014consistent}, super-resolution
\cite{wanner2014variational,wanner2012spatial,bishop2012light}
and depth estimation (see related work in Section~\ref{sec:related}). Our
depth estimation algorithm also takes advantage of this property.

\vspace{-3mm}
\section{Related work and our contributions}
\vspace{-2mm}
\label{sec:related}
Various algorithms have been proposed to estimate depth from light
field images. To construct the depth map of the sub-aperture image
$I_{s^*,t^*}$ a structure tensor on each EPI $E_{y,t^*}$
($y={1,\ldots,Y})$ and $E_{x,s^*}$ ($x={1,\ldots,X}$) is used by
\cite{wanner2012globally} to estimate the slopes of the EPI-lines. For
each pixel in the image $I_{s^*,t^*}$ they get two slope estimations,
one from each EPI. They combine both estimations by minimizing an
objective function. In that objective they use a regularization on the
gradients of the depth map to make the resulting depth map
smooth. Additionally, they encourage that object edges in the image
$I_{s^*,t^*}$ and in the depth map coincide. A coherence measure for
each slope estimation is constructed and used to combine the two slope
estimations. Despite not using the full light field but only the
sub-aperture images $I_{s^*,t}$ ($t={1,\ldots, T}$) and $I_{s,t^*}$
($s={1,\ldots S}$) for a given $(s^*,t^*)$, they achieve appealing
results.

In \cite{kim2013scene} a fast GPU-based algorithm for light fields of
about 100 high resolution DSLR images is presented. The algorithm also
makes use of the lines appearing in an EPI. The rough idea for a 3D
light field is as follows: to find the disparity of a pixel $(x,s)$ in
an EPI, its RGB pixel value is compared with all other pixel values in
the EPI that lie on a slope that includes this pixel. They loop over a
discretized range of possible slopes. If the variance in pixel color
is minimal along that slope, the slope is accepted. For a 4D light
field not a line is used, but a plane. \cite{tosic2014light} note that
uniformly colored areas in the light field form not lines but thick rays 
in the EPI's, so they convolve a Ray-Gaussian kernel with the light field
and use resulting maxima to detect areas of even depth.

In \cite{tao2013depth} the refocusing equation described in
\cite{ng2005light} is used to discretely sheer the 4D ligthfield in
order to get correspondence and defocus clues, which are combined
using a Markov random field. A similar approach is used by
\cite{diebold2013epipolar}. \cite{wang2015occlusion} build on
\cite{tao2013depth} by checking edges in the sub-aperture images for
occlusion boundaries, yielding a higher accuracy around fine
structured occlusions. \cite{lin2015depth} estimate the focal stack,
which is the set of images taken from the same position but with different
focus settings and also apply Markov random field energy minimization
for depth estimation.

\cite{zhang2015light} introduce a generative model from binary images which is initialized with a disparity map for the binary views and subsequently refined step by step in Fourier space by testing different directions for the disparity map. \cite{heber2014shape} suggest an optimization approach, which makes
use of the fact that the sub-aperture images of a light field look
very similar: sub-aperture images are warped to a reference image by
minimizing the rank of the set of warped images. In their preceding
work \cite{heber2013variational} the same authors match all
sub-aperture images against the center view using a generalized stereo
model, based on variational principles.
\cite{perwass2012single} uses a multi-focus plenoptic camera and
calculates the depth from the raw images using triangulation techniques.
\cite{adelson1992single} uses displacement estimation between
sub-aperture image pairs and combines the different estimations via a
weighted sum, where the weights are proportional to the coherence of
the displacement estimation.

\vspace{-2mm}
\paragraph{\textbf{Contributions of this paper:}}

While a number of different approaches for depth estimation from
light field images exists, we derive to the best of our knowledge for
the first time a fully generative model of EPI images that allows a
principled approach for depth map estimation and the ready
incorporation of informative priors. In particular, we show

\begin{enumerate}
\item a principled, fully generative model of light fields that
  enables the estimation of \emph{continuous} depth labels (as opposed to discrete values),
\item an efficient gradient-based optimization approach that can be
  readily extended with additional priors. In this work we demonstrate
  the integration of a powerful non-local means (NLM) prior term, and
\item favourable results on a number of challenging real-world
  examples. Especially at object boundaries, significantly sharper
  edges can be obtained.
\end{enumerate}

\newcommand{\specialcell}[2][c]{%
  \begin{tabular}[#1]{@{}c@{}}#2\end{tabular}}
\begin{figure}[t]
  \centering
  \tabcolsep0.5mm
  \scriptsize
  \begin{tabular}{cccc}
    \includegraphics[width=0.24\textwidth]{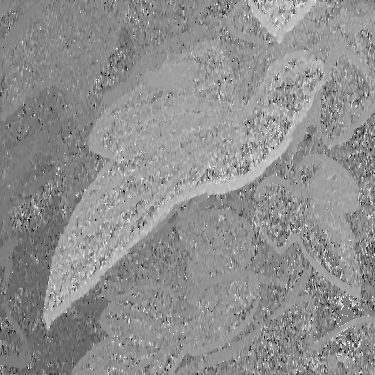}&
    \includegraphics[width=0.24\textwidth]{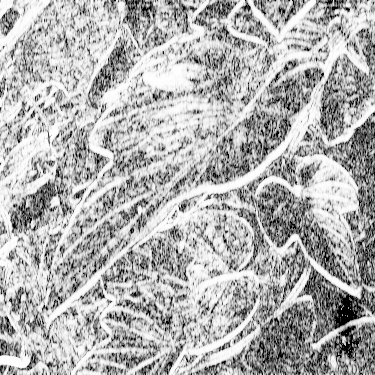}&
    \includegraphics[width=0.24\textwidth]{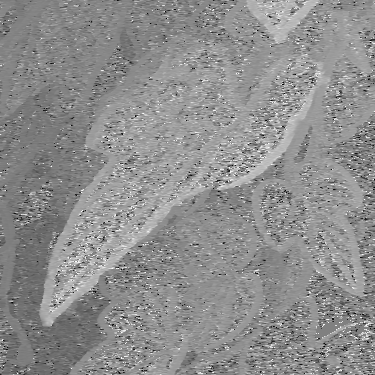}&
    \includegraphics[width=0.24\textwidth]{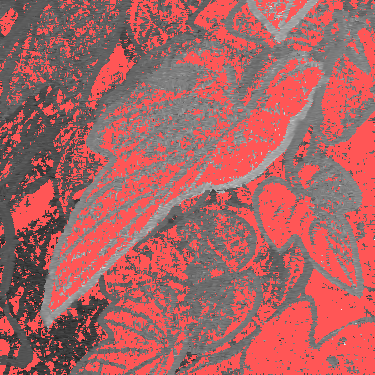}\\
    
    \specialcell{(a) depth estimation\\ on $E_{x^*,s^*}$} & 
    \specialcell{(b) coherence for depth\\ estimation on $E_{x^*,s^*}$} &
    \specialcell{(c) depth estimation\\ on $E_{y^*, t^*}$} & 
    \specialcell{(d) $m_{init}$, thresholded,\\ combined a) \& c)} \vspace{1mm} 
        \\

    \includegraphics[width=0.24\textwidth]{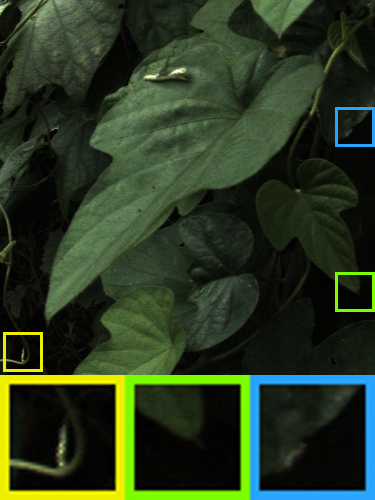}&
    \includegraphics[width=0.24\textwidth]{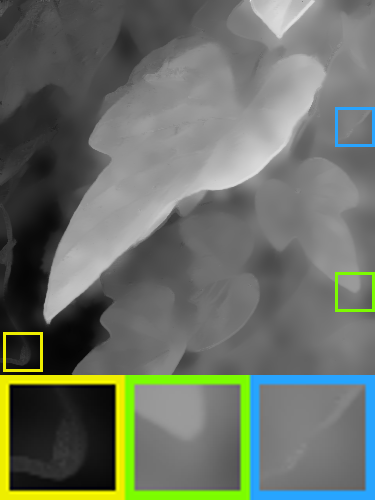} &
    \includegraphics[width=0.24\textwidth]{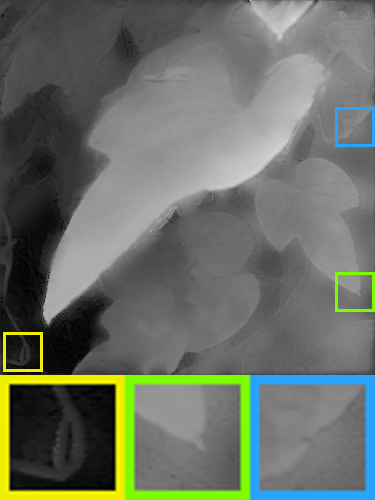} &
    \includegraphics[width=0.24\textwidth]{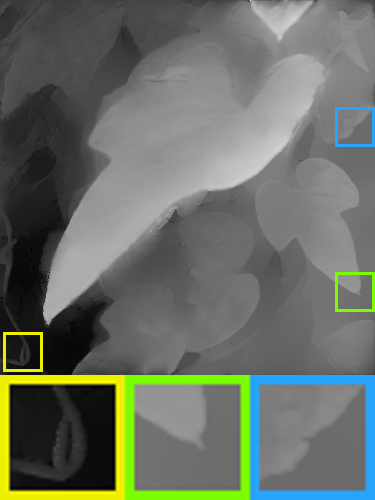}
    \\
    \specialcell{(e) center sub-aperture\\ image} & 
    \specialcell{(f) smooth propagation\\ of d)} &
    \specialcell{(g) result without\\ NLM prior} & 
    \specialcell{(h) result with\\ NLM prior} \\

  \end{tabular}                 
  \normalsize
  \caption{Overview of our method on an image by \cite{tao2013depth},
    best viewed on screen. Note: each image was normalized to [0,1],
    so gray values between images may vary. Images (a)\&(c) show rough
    depth maps, computed using the first part of the local depth
    estimation method of \cite{wanner2012globally}, (a) on the EPI
    $E_{x^*,s^*}$, and (c) on the EPI $E_{y^*, t^*}$. (b) Coherence
    map for the depth estimation on the EPI $E_{x^*,s^*}$. (d) The two
    noisy depth maps are thresholded using their corresponding
    coherence maps and then combined.  Evidently, coherent depth
    values are mainly located at edges while non-coherent depth values
    (in red) occur in smoother regions. (f) To fill the non-coherent
    areas, the coherent depth values are propagated using an NLM
    approach. (g) Our result without a prior shows more details than
    (f), but also introduces speckles in the depth map. (h) The
    NLM prior is able to get reduce speckles while preserving fine
    details, especially on edges.}
  \label{fig:overview}
\end{figure}

\vspace{-6mm}
\section{Overview}
\label{sec:overview}
\vspace{-3mm}
Our proposed algorithm computes the depth map of a selected
sub-aperture image from a light field. In this paper -- without loss
of generality -- we choose the center sub-aperture image. Our method
works in a two step procedure: as a first step, a rough estimation of
the depth map is computed locally
(Sec.~\ref{sec:rough_depth_estimate}). This serves as an
initialization to the second step (Sec.~\ref{sec:refinement_step}), in
which the estimated depth map is refined using a gradient based
optimization approach.

The following implicit assumptions are made by our algorithm: 
\begin{itemize}
\item All objects in the scene are Lambertian, a common assumption,
  which is also made by
  \cite{wanner2014variational,sebe2002multi,kim2013scene,chai2000plenoptic}.
\item There exist no occlusions in the scene. This means that lines in
  the EPI images do not cross. This is a reasonable assumption,
  especially for the Lytro camera, as the angular resolution is not
  very large, as also stated by
  \cite{dansereau2011plenoptic,dansereau2013light}.\footnote{We
    discuss the implication of the no-occlusion assumption in more
    detail in the supplemental material.}
\item Object boundaries in the depth image coincide with changes in the RGB image. This assumption is also explicitly
  made in \cite{park2011high,ferstl2013image}.
\end{itemize}
Note that these assumptions are not perfectly met in real scenes. Despite
relying implicitly on these assumptions, our algorithm works well on
real world images as we will demonstrate in Sec.~\ref{sec:exper-comp}.
\vspace{-3mm}
\section{Rough depth map estimation using NLM}
\label{sec:rough_depth_estimate}
\vspace{-1mm}
To get an initialization for the depth map, we adopt the initial part
of the depth estimation algorithm of
\cite{wanner2012globally}\footnote{We use the given default values
  $\sigma=1.0$ and $\tau=0.5$.}: the local depth estimation on
EPIs. It estimates the slopes of the lines visible in the epipolar
image using the structure tensor of the EPI. It only uses the epipolar
images of the angular center row ($E_{y,t^*}$ with $t^* = T/2$,
$y=\{1,\ldots Y\}$) and angular center column ($E_{x,s^*}$ with $s^* =
S/2$, $x=\{1,\ldots X \}$) and yields two rough depth estimates for
each pixel in the center image, see Fig.~\ref{fig:overview} (a,c).
Additionally, it returns a coherence map $C(\cdot)$ that provides a
measure for the certainty of the estimated depth values
(Fig.~\ref{fig:overview} (b)).

This approach returns consistent estimates at image edges but
provides quite noisy and unreliable estimates at smoother regions, see
Fig.~\ref{fig:overview} (a,c). We take the two noisy depth estimates
and threshold them by setting depth values with a low coherence
value to \emph{non-defined}. This gives us an initial estimate for the
depth values at the edges. After thresholding, the two estimates are
combined: if both estimates on $E_{y,t^*}$ and $E_{x,s^*}$ for a
pixel in the depth map have coherent values, we will take the
depth estimation with the higher coherence value. The result is shown
in Fig.~\ref{fig:overview} (d). This rough depth map estimate is
denoted by $m_{init}$.

To propagate depth information into regions of non-coherent pixels (red
pixels in Fig.~\ref{fig:overview} (d)), we solve the optimization problem
\begin{equation}
  \label{eq:structTensor}
  \operatornamewithlimits{argmin}_m  \sum\limits_p \sum\limits_{q
    \in N(p)}  w_{pq} \biggl[  \,  \, \big( m(p) - m(q) \big)^2 
   + \,  C(p) \, \big(m_{init}(p) - m(q)\big)^2 
  \biggr]
\end{equation}
where $w_{pq}\in[0,1]$ is a weighting term on the RGB image that
captures the similarity of the color and gradient values of the $3
\times 3$ window around $p$ and $q$ (see
Sec.~\ref{sec:NLM_regularizer}, Eq.~\eqref{eq:NLM_wpq}).  $N(p)$ are
the neighboring pixels around $p$, e.g. in an $11 \times 11$ window
with $p$ in the center.  The term $w_{pq}( m(p) - m(q))^2$ ensures
that pixels in $N(p)$ have similar depth values as $p$ if they are
similar in the RGB image (high $w_{pq}$).  The term $w_{pq} C(p) (
m(q) - m_{init}(p) )^2$ enforces that a pixel with high coherence
$C(p)$ propagates its depth value $m_{init}(p)$ to neighboring pixels
$q$ which are similar in the RGB image. Fig.~\ref{fig:overview}(f)
shows the result of this propagation step. Eq.~\eqref{eq:structTensor}
is minimized using {L-BFGS-B} \cite{lbfgsb_matlab_wrapper}.

\begin{figure}[tb]
  \centering
  \tabcolsep0.5mm
  \subfloat[Overview of the refinement step visualized with
  sub-aperture images, best viewed on screen: The estimated light
  field (LF) is constructed by shifting pixels from the center
  sub-aperture image to the other sub-aperture images.
  The shifting depends on the current depth value of the shifted
  pixel and the distance to the sub-aperture image.
  The objective is to minimize the
  squared L2 distance between the estimated LF and the original LF.
  ]{%
     \includegraphics[width=0.99\textwidth]{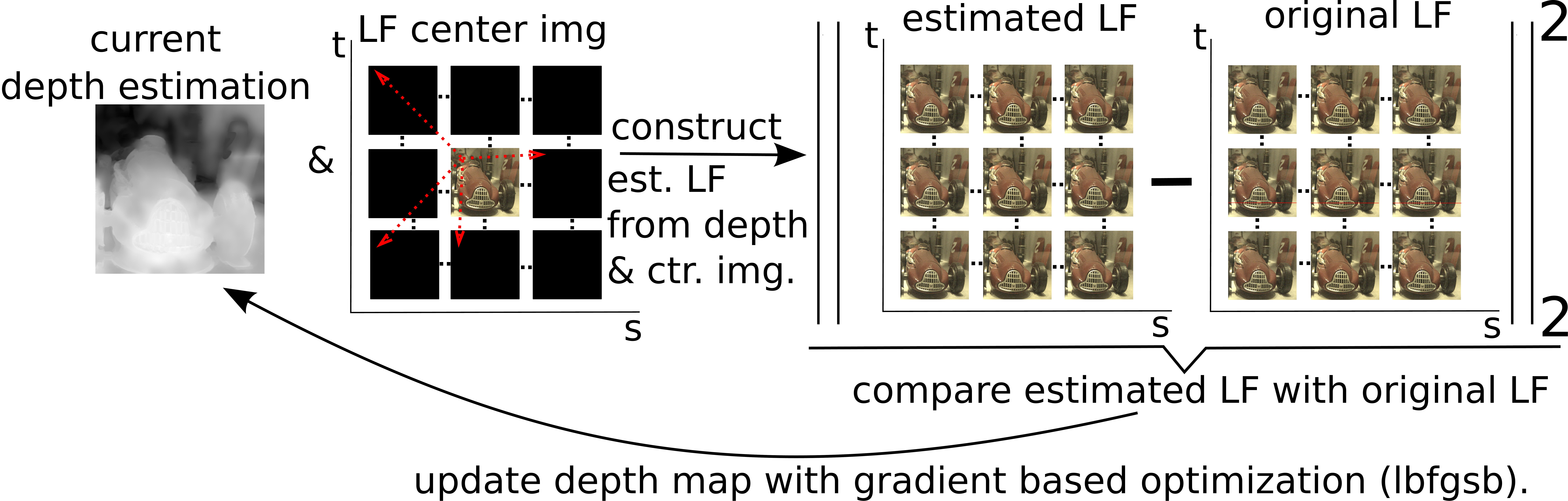} 
    \label{fig:overview_refinement}    
  }
  
  \subfloat[{\bf Top}: Original EPI $E_{y^*,t^*}$ from 7 sub-aperture images
  using the row in each image, which is highlighted in red
  above. {\bf Bottom}: Estimated EPI constructed solely from the center row by
  shifting center row pixels according to their depth values.
  Note that for better visualization, the EPI's are vertically
  stretched to twice their heights. Four red lines are
  overlaid to visualize the emergence of lines in the EPI. ]{%
    \includegraphics[width=0.99\textwidth]{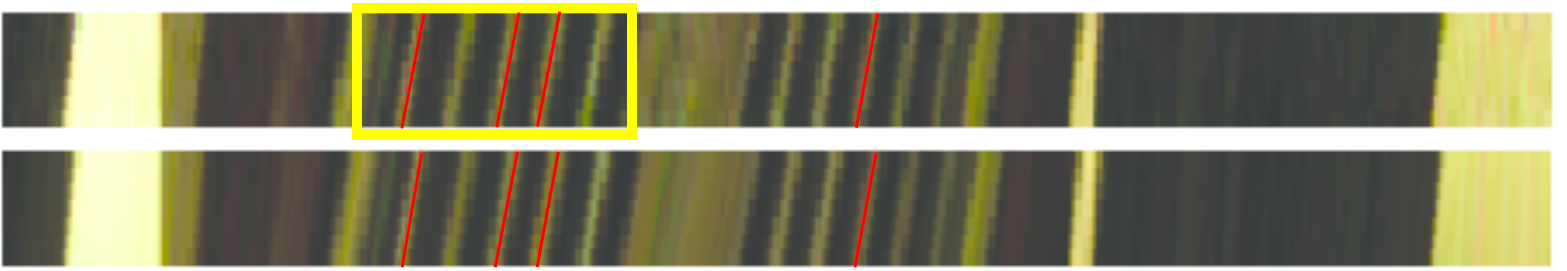} 
    \label{fig:overview_refinement_epi}    
  }

  \subfloat[Overview of the refinement step visualized with an EPI.\vspace{-3mm}
]{%
    \includegraphics[width=0.90\textwidth]{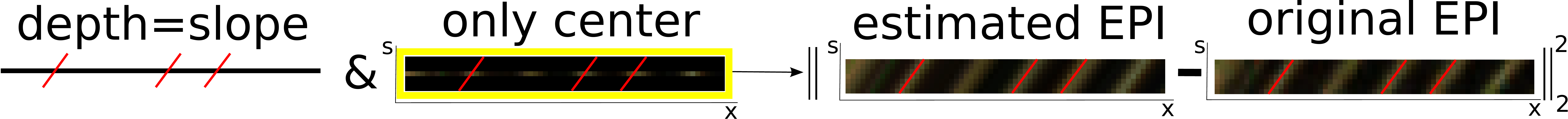} 
    \label{fig:overview_refinement_epidepth}    
  }
  \caption{Overview of the refinement step.}
\end{figure}
\vspace{-3mm}
\section{Refinement step}
\vspace{-1mm}
\label{sec:refinement_step}
The refinement step is based on the observation that the sub-aperture
images can be almost entirely explained and predicted from the center
image $I:=I_{S/2,T/2}$ alone, namely by shifting the pixels from the
image $I$ along the lines visible in the epipolar image. This has
already been observed by others
\cite{gortler1996lumigraph,isaksen2000dynamically,zhang2015light}.
Fig.~\ref{fig:overview_refinement_epidepth} visualizes the case for an
EPI $E_{y^*,t^*}$.

To make full use of the recorded 4D light field, the center image
pixels are not only moved along a line, but along a 2D plane. For a
fixed $s^*$ or $t^*$ this 2D plane becomes a line visible in the
respective EPI views $E_{x^*,s^*}$ or $E_{y^*,t^*}$. Note that the slopes
of the line $m_{x_c,y_c}$ going through the center pixel $(x_c,y_c)$
are the same in $E_{y_c,t^*}$ and $E_{x_c,s^*}$.

When moving a center image pixel $I(x_c,y_c)$ along the 2D plane with
given slope $m_{x_c,y_c}$, its new coordinates in the sub-aperture
image $I_{S/2+d_s, T/2 + d_t}$ become $(x_c + m_{x_c,y_c} d_s, y_c +
m_{x_c,y_c} d_t)$, where $d_s$ and $d_t$ denote the angular distances
from the center sub-aperture image to the target sub-aperture image,
formally ${d_s = s - S/2}$, ${d_t = t - T/2}$. The estimated light field
that arises from shifting pixels from the center sub-aperture image to
all other sub-aperture views is denoted by $\widetilde{L}_m(s,t,x,y)$.

\begin{figure}[h]
  \centering
  \includegraphics[width=0.49\textwidth]{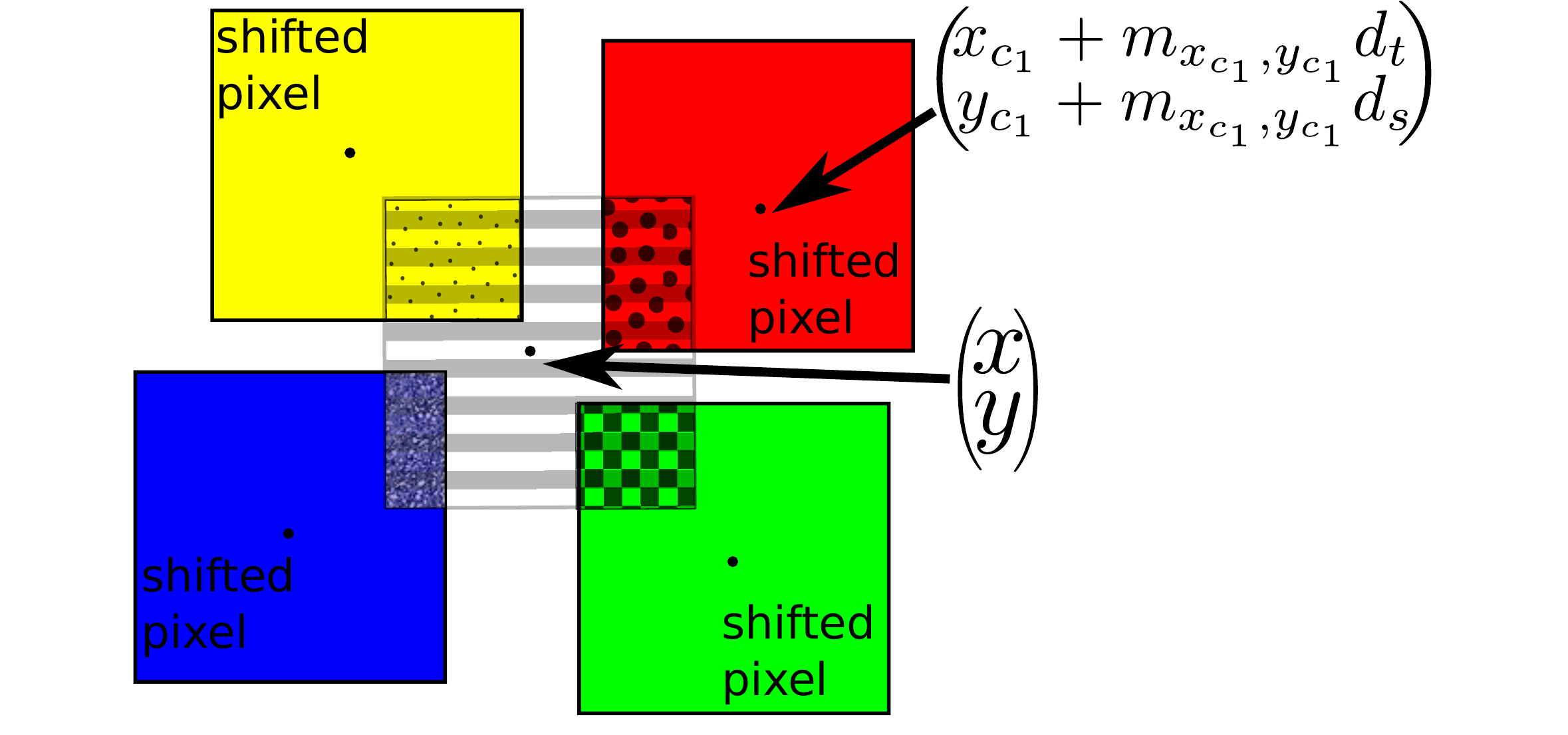}
  \includegraphics[width=0.49\textwidth]{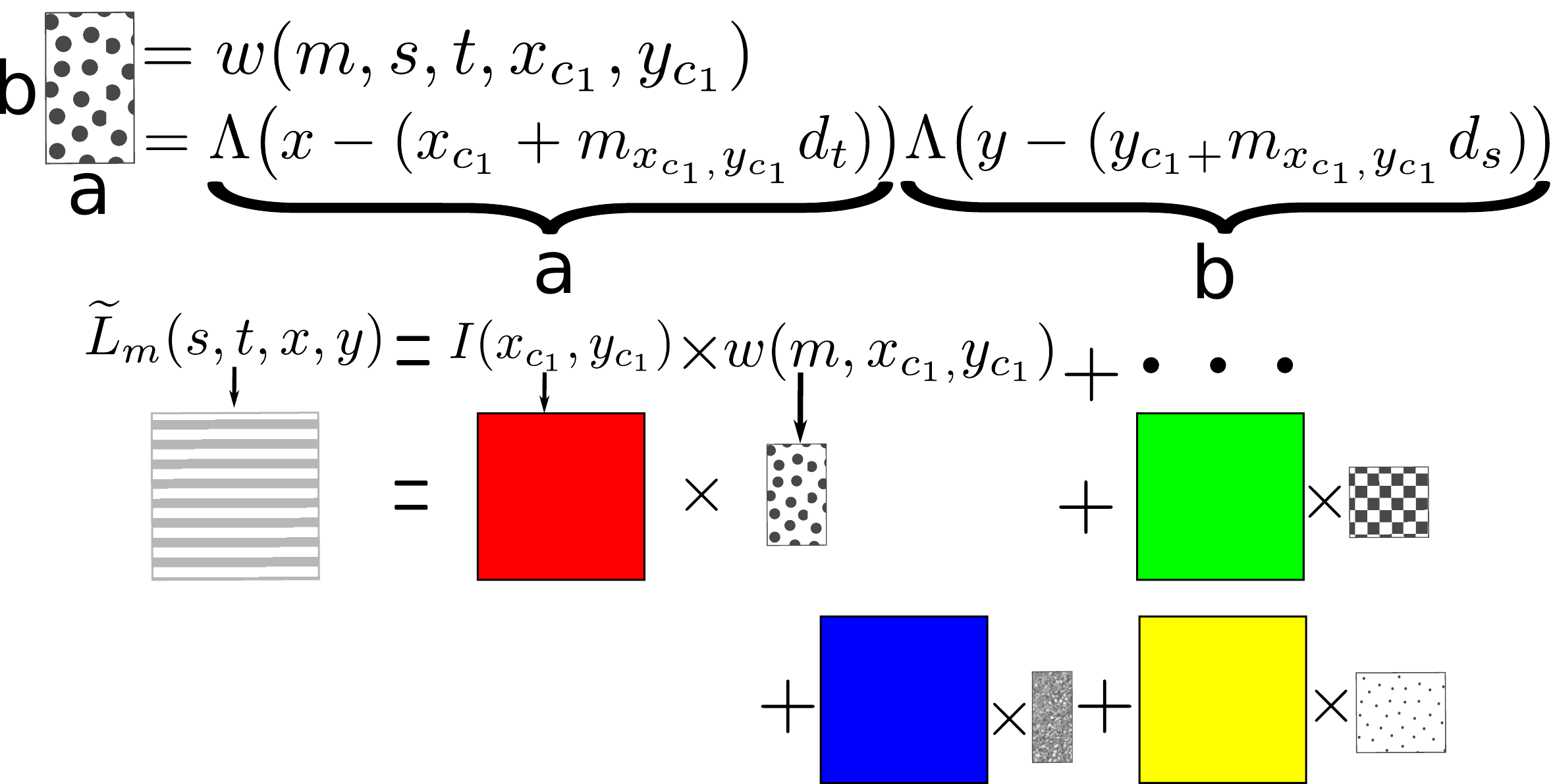}
  \caption{Modified Bilinear Interpolation: The color of the estimated pixel is
    determined by using a weighted sum of the shifted pixels that
    overlap with it. The weight of each pixel is the area of overlap
    with the estimated pixel.}
  \label{fig:bilinearInterpolation_explained}
\end{figure}

The influence of pixel $I(x_c,y_c)$ on pixel
$\widetilde{L}_m(s,t,x,y)$ is determined by the distances $x-(x_c +
m_{x_c,y_c}d_s)$ and $y- (y_c + m_{x_c,y_c} d_t)$. Only if the
absolute values of both distances are smaller than one, the pixel
$I(x_c,y_c)$ influences $\widetilde{L}_m(s,t,x,y)$, see
Fig.~\ref{fig:bilinearInterpolation_explained}. Any pixel in the light
field $\widetilde{L}_m(s,t,x,y)$ can thus be computed as
the weighted sum of all pixels from the center image $I$:
\begin{equation*}
  \label{eq:data-term}
  \widetilde{L}_m(s,t,x,y)
  =\sum\limits_{x_c} \sum\limits_{y_c} I(x_c,y_c)
   \cdot \Lambda\big(x-(x_c+m_{x_c,y_c}d_t) \big)
   \cdot \Lambda\big(y-(y_c+m_{x_c,y_c}d_s) \big)
\end{equation*}
where $\Lambda: \mathbb{R} \to [0,1]$ is a weighting function and
denotes the differentiable version 
of the triangular function defined as
\begin{equation*}
  \bar{\Lambda}(x) =
  \begin{cases}
    1+x & \mbox{if } -1 < x < 0 \\
    1-x & \mbox{if } 0 \le x < 1 \\
    0   & \mbox{otherwise, i.e. if } |x| \ge 1
  \end{cases}
\end{equation*}
\vspace{-3mm}
\vspace{-4mm}
To allow for sub-pixel shifts, we use the following bilinear
interpolation scheme: As shown in Fig.~\ref{fig:bilinearInterpolation_explained},
the intensity value of the pixel $\widetilde{L}_m(s,t,x,y)$ is the
weighted sum of all overlapping pixels (the four colored
pixels in Fig.~\ref{fig:bilinearInterpolation_explained}).
The weight of each pixel is the area of overlap with the
pixel $\widetilde{L}_m(s,t,x,y)$, which is given by the triangular term.

Note that in practice, it is not necessary to iterate over all pixels in the center image
for every pixel $\widetilde{L}_m(s,t,x,y)$. Instead, the projection of each pixel in the center image
is calculated and added to the (up to) 4 pixels in the subview where the overlapping area is nonzero,
leading to a linear runtime complexity in the size of the light field.
To get a refined depth estimation map we optimize the following
objective function:
\begin{equation*}
  \label{eq:Refinement_objective}
  m = \operatornamewithlimits{argmin}_m || \widetilde{L}_m - L ||^2_2 +
  \lambda \,R(m)
\end{equation*}
where $R(m)$ is a regularization term which we will describe in more detail
in the remainder of this section.

\subsection{Non-local means regularization}
\label{sec:NLM_regularizer}
The NLM regularizer was first proposed in \cite{buades2005non} and has proven useful for 
the refinement of depth maps in other contexts \cite{favaro2010recovering}.
We define it as
\begin{equation*}
  R(m) = \sum\limits_p \sum\limits_{q \in N(p)} w_{pq} (m(p) - m(q))^2
\end{equation*}
with $N(p)$ being the search window around a pixel $p$, e.g. an $11
\times 11$ window, and $w_{pq}$ is the weight expressing the similarity
of the pixels $p$ and $q$.  We define $w_{pq}$ as
\begin{equation}
  w_{pq} = \exp\Bigg(-\mathop{\sum_{p' \in N'(p)}}_{q' \in N'(q)}
  \frac{[I(p') -  I(q')]^2}{\sigma_{Color}^2} 
  + \frac{[\nabla_{p,p'}I - \nabla_{q,q'} I ]^2}{\sigma_{Grad}^2} \Bigg)
\label{eq:NLM_wpq}
\end{equation}
where $\sigma_{Color}^2$ and $\sigma_{Grad}^2$ are the variances
in the color and gradient values of the image and
$\nabla_{p,p'} I := I(p) - I(p')$ is the image gradient at pixel
position $p$. This encourages edges in the depth map at locations
where edges exist in the RGB image and smoothens out the remaining
regions. In all experiments, $N'$ was a $3 \times 3$ window.

 \section{Implementation Details}

 \label{sec:optimization}
 
 Our implementation is in MATLAB. For numerical optimization we used
 the MATLAB interface by Peter Carbonetto \cite{lbfgsb_matlab_wrapper}
 of the gradient based optimizer {L-BFGS-B} \cite{zhu1997algorithm}.
 The only code-wise optimization we applied is implementing some of
 the computationally expensive parts (light field synthesis) in C~(MEX)
 using the multiprocessing API OpenMP\footnote{\url{http://openmp.org}}.
 Current runtime for computing the depth
 map from the $7 \times 7 \times 375 \times 375 \times 3$ light field
 from the Lytro camera is about 270 seconds on a 64bit
 Intel Xeon CPU E5-2650L 0 @ 1.80GHz architecture using 8 cores.
 All parameter settings for the experiments are included in
 the supplementary and in the source code and are omitted here for brevity.
 In particular, we refer the interested reader to the supplementary for
 explanations of their influence to the depth map estimation along with
 reasonable ranges that work well for different types of images.
 The code as well as our Lytro dataset is publicly available.\footnote{\url{http://webdav.tue.mpg.de/pixel/lightfield_depth_estimation/}}

\vspace{-2mm}
\section{Experimental results}
\label{sec:exper-comp}
\vspace{-2mm}
\subsection*{Comparison on Lytro images}
\vspace{-2mm}
In Fig.~\ref{fig:comparison_on_Tao} we show a comprehensive comparison
of our results with several other works including the recent work of
\cite{wang2015occlusion} and \cite{lin2015depth}. The resolution of
the Lytro light field is $S\times T = 7 \times 7$,
and the resolution of each sub-aperture image is $X \times Y = 375 \times 375$
pixels.
We decoded the raw Lytro images with the algorithm by
\cite{dansereau2013decoding}\footnote{The images have neither been gamma
  compressed nor rectified.}, whereas others have developed proprietary decoding procedures leading to slightly different image dimensions.
  
Overall, our algorithm is able to recover more details and better
defined edges while introducing no speckles to the depth maps. Depth
variations on surfaces are much smoother in our results than
e.g. in \cite{wang2015occlusion} or \cite{lin2015depth} while being able
to preserve sharp edges. Even small details are resolved in the depth
map, see e.g. the small petiole in the lower left corner of the leaf
image (second row from the top, not visible in print).

\newcommand \taoimgsize {0.1405}
\begin{figure}[tb]
  \centering
   \tabcolsep0.5mm
  \begin{tabular}{ccccccc}
  
      Reference &
      Our results &
      \cite{tao2013depth} &
      \cite{wang2015occlusion} &
      \cite{lin2015depth} &
      \cite{sun2010secrets} &
      \cite{wanner2012globally} \\

	  \includegraphics[height=\taoimgsize\textwidth]{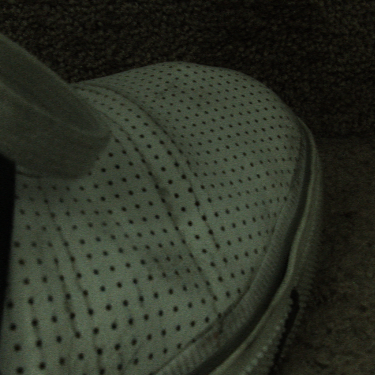} &
	  \includegraphics[height=\taoimgsize\textwidth]{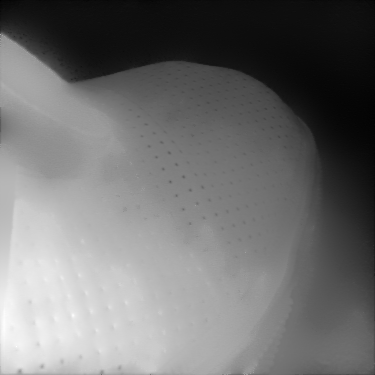} &
	  \includegraphics[height=\taoimgsize\textwidth]{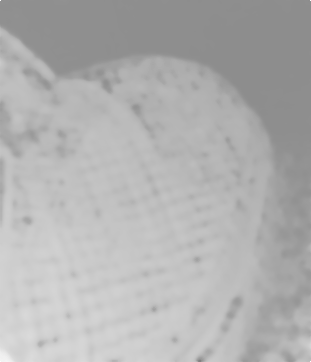} &
	  \includegraphics[height=\taoimgsize\textwidth]{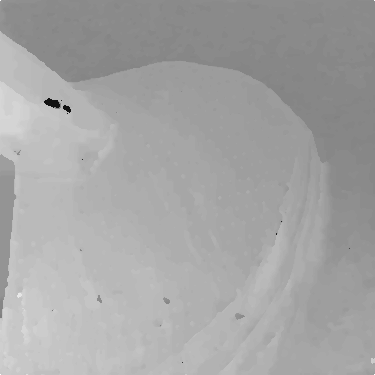} &
	  \includegraphics[height=\taoimgsize\textwidth]{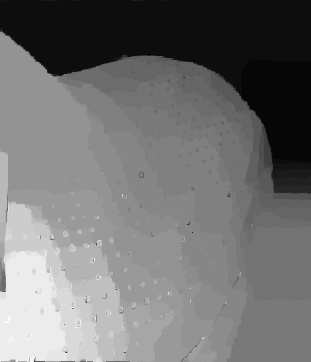} &
	  \includegraphics[height=\taoimgsize\textwidth]{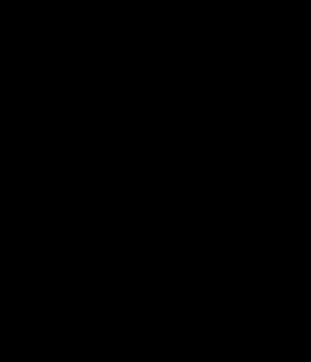} &
	  \includegraphics[height=\taoimgsize\textwidth]{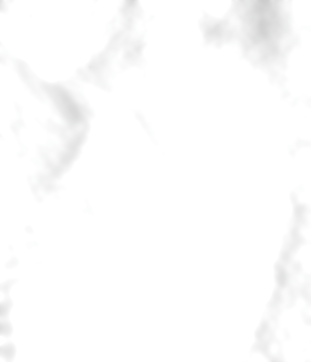} \\

    \includegraphics[height=\taoimgsize\textwidth]{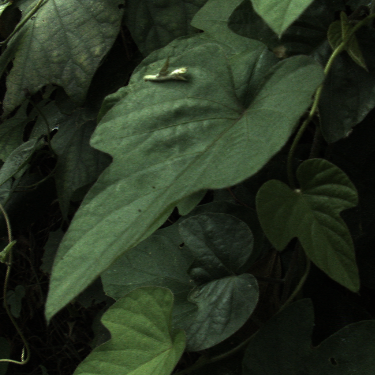} &
    \includegraphics[height=\taoimgsize\textwidth]{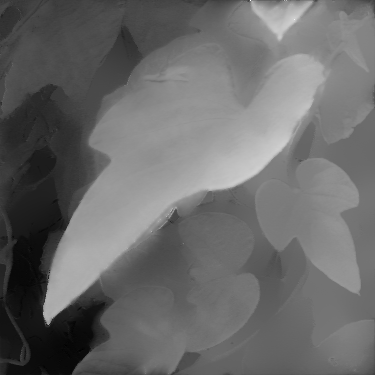} &
    \includegraphics[height=\taoimgsize\textwidth]{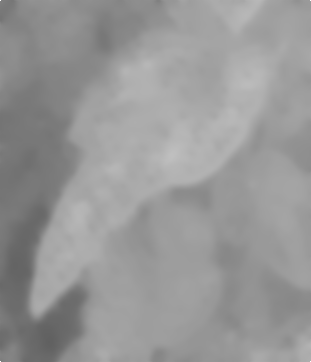} & 
    \includegraphics[height=\taoimgsize\textwidth]{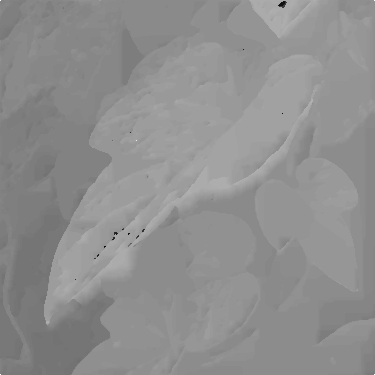} &
    \includegraphics[height=\taoimgsize\textwidth]{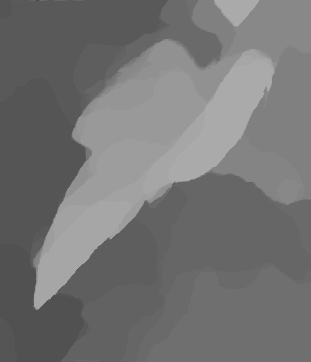} &
    \includegraphics[height=\taoimgsize\textwidth]{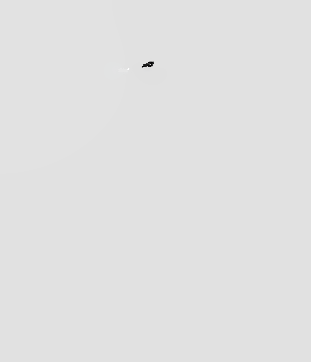} &
    \includegraphics[height=\taoimgsize\textwidth]{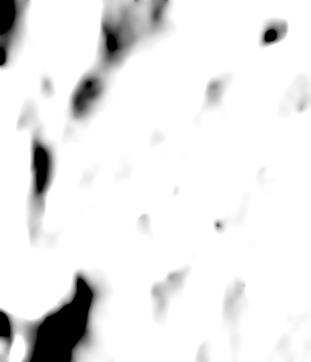} \\
    
    \includegraphics[height=\taoimgsize\textwidth]{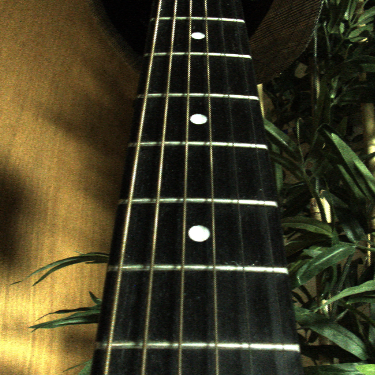} &
    \includegraphics[height=\taoimgsize\textwidth]{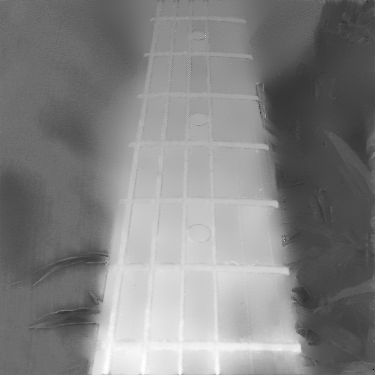} &
    \includegraphics[height=\taoimgsize\textwidth]{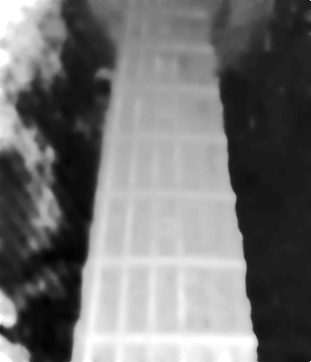} &
    \includegraphics[height=\taoimgsize\textwidth]{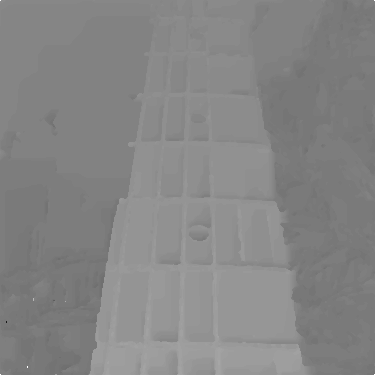} &
    \includegraphics[height=\taoimgsize\textwidth]{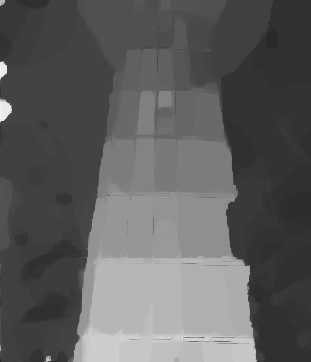} &
    \includegraphics[height=\taoimgsize\textwidth]{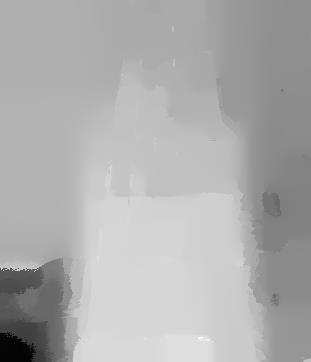} &
    \includegraphics[height=\taoimgsize\textwidth]{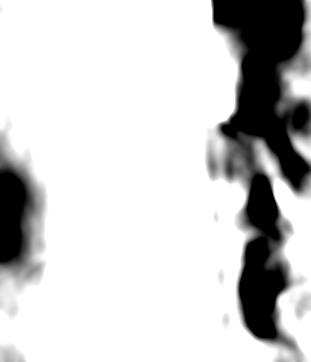} \\
    
    \includegraphics[height=\taoimgsize\textwidth]{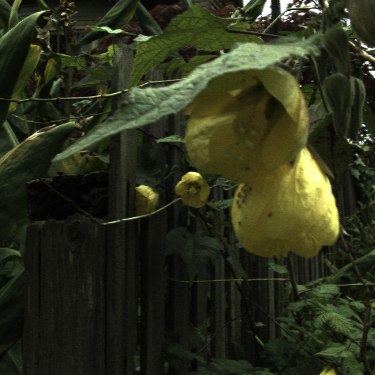}      &
    \includegraphics[height=\taoimgsize\textwidth]{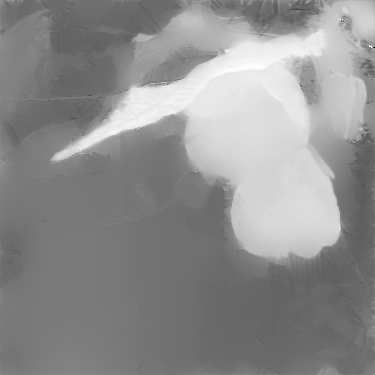} &
    \includegraphics[height=\taoimgsize\textwidth]{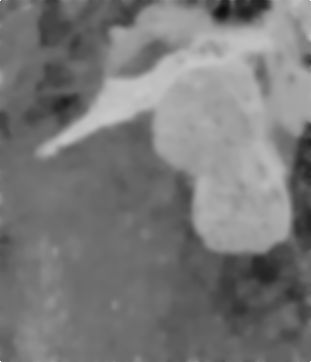} &
    \includegraphics[height=\taoimgsize\textwidth]{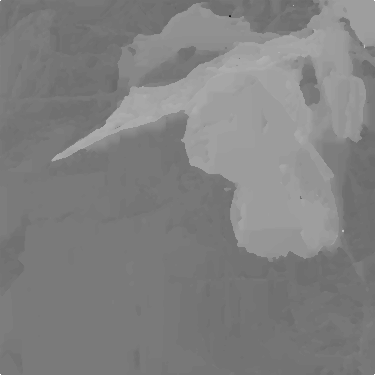} &
    \includegraphics[height=\taoimgsize\textwidth]{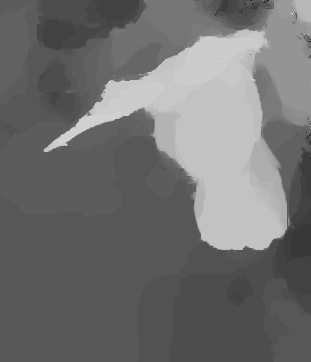} &
    \includegraphics[height=\taoimgsize\textwidth]{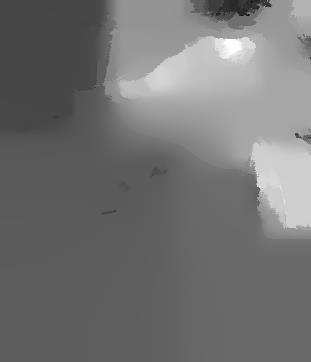} &
    \includegraphics[height=\taoimgsize\textwidth]{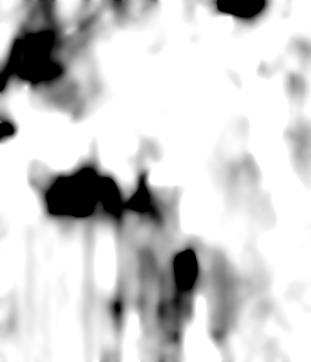}

 \end{tabular}
  \caption{Comparison on Lytro images taken from \cite{tao2013depth}, best
    viewed on screen.
    The results for \cite{sun2010secrets}
    and \cite{wanner2012globally} are taken from \cite{lin2015depth}.
    Our algorithm is able to recover finer details
    and produces fewer speckles in the depth map. Note that \cite{sun2010secrets} is originally an optical flow algorithm whose results are included for comparison.}
  \label{fig:comparison_on_Tao}
\end{figure}

\vspace{-2mm}
\subsection*{Comparison on the Stanford light field dataset}
\vspace{-2mm}
\begin{figure}[b!t]
  \centering
  \tabcolsep0.7mm
  \begin{tabular}{ccccc}
    \includegraphics[width=0.23\textwidth]{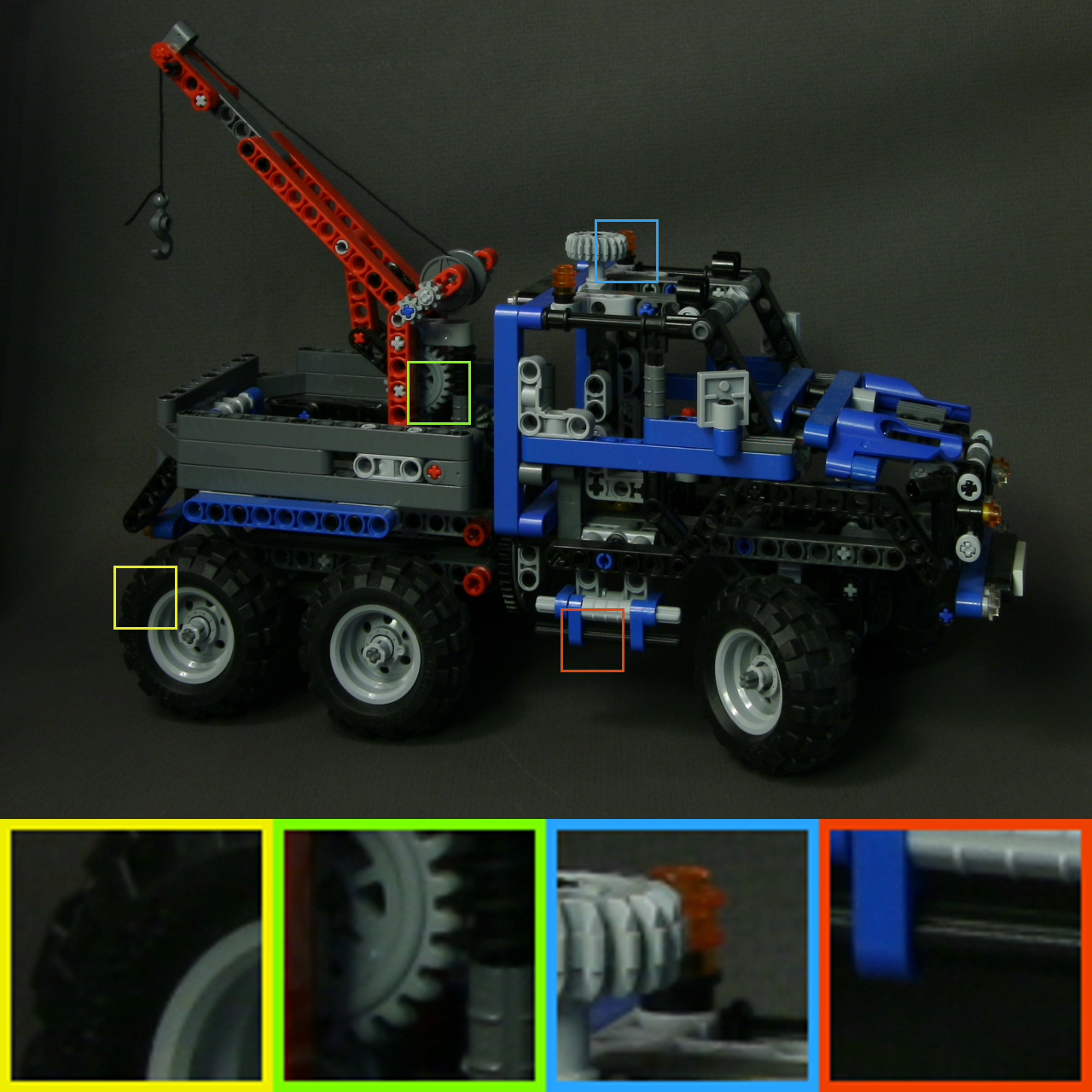} &
    \includegraphics[width=0.23\textwidth]{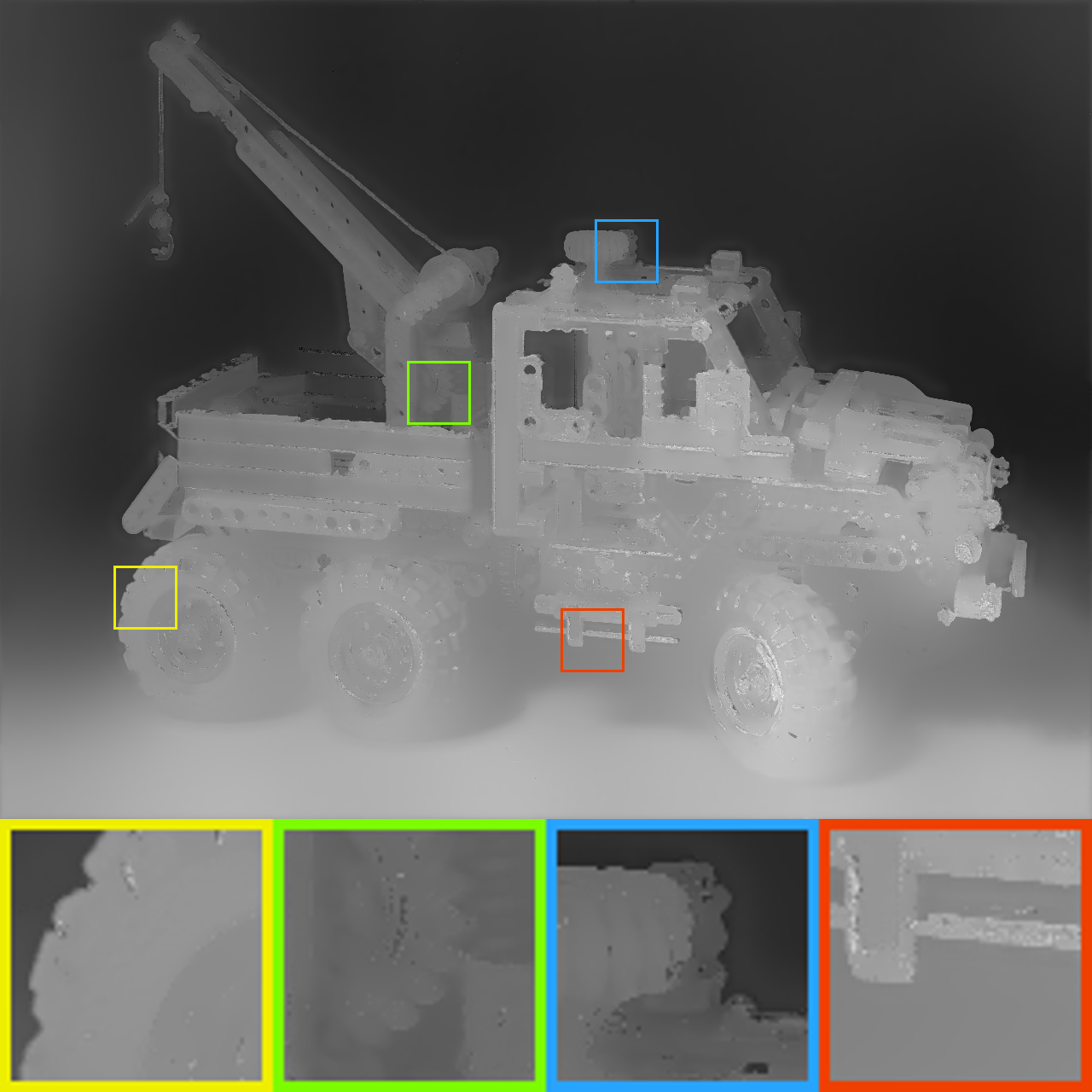}&
    \includegraphics[width=0.23\textwidth]{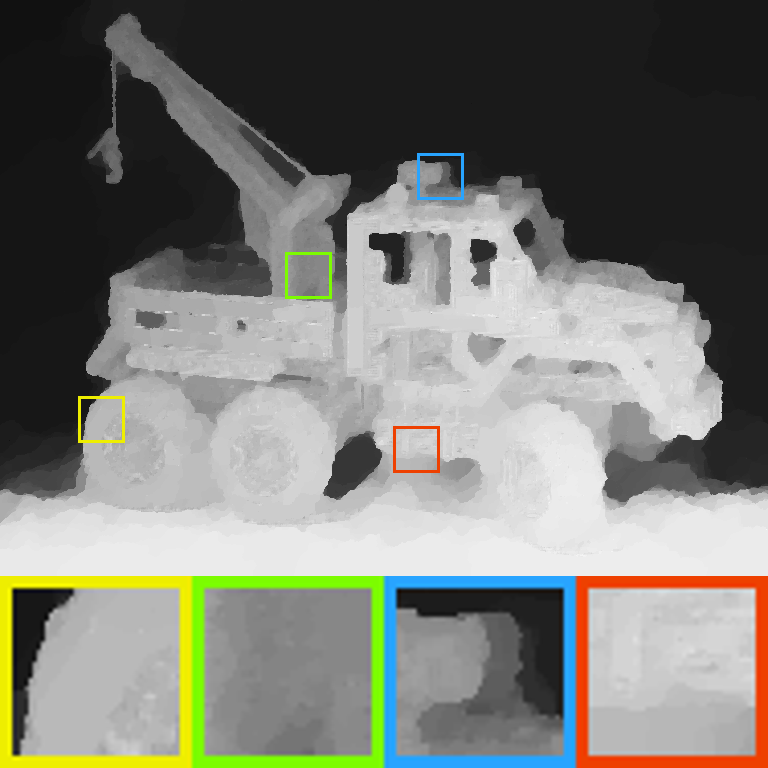} &
    \includegraphics[width=0.23\textwidth]{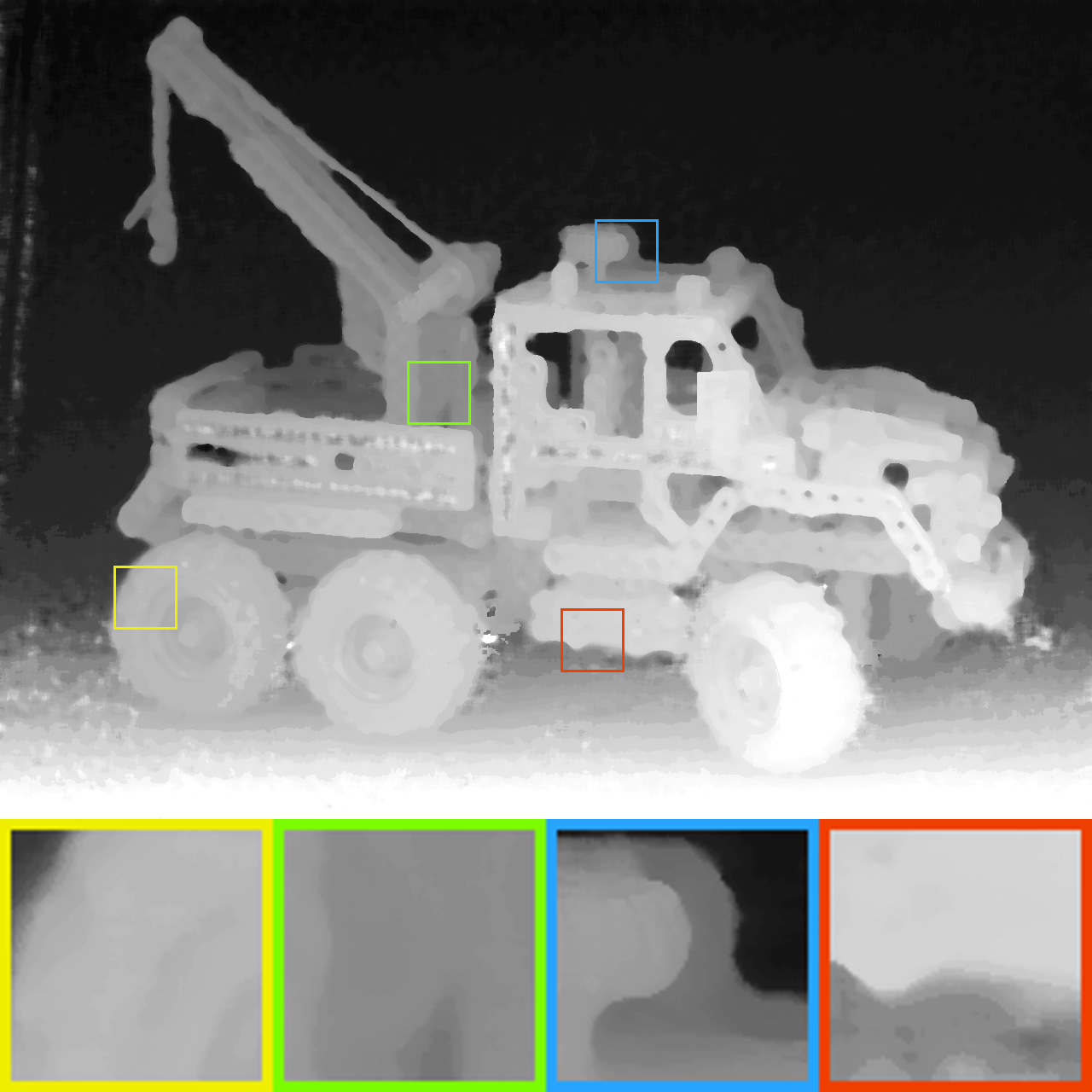}\\
    (a) center image  & 
    (b) our result  & 
    (c) result of \cite{wanner2012globally}  & 
    (d) result of \cite{kim2013scene} \\    
  \end{tabular}

  \caption{Comparison on images of the Stanford light
      field dataset \cite{stanfordLFdataset}, best viewed on
      screen. We are able to recover finer details and our depth
      boundaries match better with the RGB image boundaries. The resolution is $1280
      \times 960$ for all images except for \cite{wanner2012globally}, where it is  $768 \times 576$
      as the image is taken from the respective paper.}
  \label{fig:result_stanford_truck}
\normalsize
\end{figure}

In Fig.~\ref{fig:result_stanford_truck} we compare our method on the
truck image of the Stanford light field dataset \cite{stanfordLFdataset}. The
light field has a resolution of $S \times T = 17 \times 17$ sub-aperture views.
Each sub-aperture image has a resolution of $X
\times Y =1280 \times 960$ pixels. Compared to the already visually
pleasing results of \cite{wanner2012globally} and \cite{kim2013scene},
we are able to recover finer and more accurate details (see closeups
in Fig.~\ref{fig:result_stanford_truck}). Additionally, the edges of
the depth map match better with the edges of the RGB image. The results
on the Amethyst image and further comparisons are shown in the supplementary.

\subsection*{Quantitative evaluation on artificial dataset}
We compare our algorithm with the state of the art \cite{wang2015occlusion} on
the artificial dataset provided by \cite{wanner2013datasets}. We could not reproduce the
given numerical results, so we used the provided code for the comparison.
The average root mean square error on the whole
dataset\footnote{Buddha, buddha2, horses, medieval, monasRoom, papillon, stillLife.}
is \textbf{0.067} for \cite{wang2015occlusion} and
\textbf{0.063} for our algoritm. The depth maps and further details are given in the supplementary.

\section{Conclusion and future work}
We presented a novel approach to estimate the depth map from light
field images. A crucial ingredient of our approach is a generative model
for light field images which can also be used for other image processing
tasks on light fields. Our approach consists of two steps. First, a
rough initialization of the depth map is computed. In the second step,
this initialization is refined by using a gradient based optimization
approach.

We have evaluated our approach on a light field image from the
Stanford dataset \cite{stanfordLFdataset}, on real-world images
taken with a Lytro camera and on artificially generated light field images.
Despite the Lytro images being rather
noisy, our recovered depth maps exhibit fine details with well-defined
boundaries.

Our work can be extended and improved in several directions. The
algorithm lends itself well to parallelization and seems ideally
suited for a GPU implementation. Another interesting direction is to
modify our generative light field model to explicitly account for
occlusions. While our forward model can be readily adapted to allow
for cross-sections in the EPI, optimization becomes more
intricate. We leave a solution to this problem for future work.
\newpage

\bibliographystyle{splncs03}
\bibliography{gcpr2016submission_clean}

\begin{thebibliography}{10}
\providecommand{\url}[1]{\texttt{#1}}
\providecommand{\urlprefix}{URL }

\bibitem{stanfordLFdataset}
The (new) stanford light field archive. http://lightfield.stanford.edu (2008),
  [Online; accessed 07-April-2016]

\bibitem{adelson1992single}
Adelson, E.H., Wang, J.Y.A.: Single lens stereo with a plenoptic camera. IEEE
  Transactions on Pattern Analysis and Machine Intelligence (PAMI)  14(2),
  99--106 (1992)

\bibitem{bishop2012light}
Bishop, T.E., Favaro, P.: The light field camera: Extended depth of field,
  aliasing, and superresolution. IEEE Transactions on Pattern Analysis and
  Machine Intelligence (PAMI)  34(5),  972--986 (2012)

\bibitem{bolles1987epipolar}
Bolles, R.C., Baker, H.H., Marimont, D.H.: Epipolar-plane image analysis: An
  approach to determining structure from motion. International Journal of
  Computer Vision  1(1),  7--55 (1987)

\bibitem{buades2005non}
Buades, A., Coll, B., Morel, J.M.: A non-local algorithm for image denoising.
  In: CVPR (2005)

\bibitem{lbfgsb_matlab_wrapper}
Carbonetto, P.: A programming interface for {L-BFGS-B} in {MATLAB}.
  https://github.com/pcarbo/lbfgsb-matlab (2014), [Online; accessed
  15-April-2015]

\bibitem{chai2000plenoptic}
Chai, J.X., Tong, X., Chan, S.C., Shum, H.Y.: Plenoptic sampling. In: ACM
  SIGGRAPH (2000)

\bibitem{cho2014consistent}
Cho, D., Kim, S., Tai, Y.W.: Consistent matting for light field images. In:
  ECCV (2014)

\bibitem{dansereau2013light}
Dansereau, D.G., Bongiorno, D.L., Pizarro, O., Williams, S.B.: Light field
  image denoising using a linear {4D} frequency-hyperfan all-in-focus filter.
  In: IS\&T/SPIE Electronic Imaging (2013)

\bibitem{dansereau2011plenoptic}
Dansereau, D.G., Mahon, I., Pizarro, O., Williams, S.B.: Plenoptic flow:
  Closed-form visual odometry for light field cameras. In: IROS (2011)

\bibitem{dansereau2013decoding}
Dansereau, D.G., Pizarro, O., Williams, S.B.: Decoding, calibration and
  rectification for lenselet-based plenoptic cameras. In: CVPR (2013)

\bibitem{diebold2013epipolar}
Diebold, M., Goldl{\"u}cke, B.: Epipolar plane image refocusing for improved
  depth estimation and occlusion handling. In: Annual Workshop on Vision,
  Modeling and Visualization: VMV (2013)

\bibitem{favaro2010recovering}
Favaro, P.: Recovering thin structures via nonlocal-means regularization with
  application to depth from defocus. In: CVPR (2010)

\bibitem{ferstl2013image}
Ferstl, D., Reinbacher, C., Ranftl, R., R{\"u}ther, M., Bischof, H.: Image
  guided depth upsampling using anisotropic total generalized variation. In:
  ICCV (2013)

\bibitem{goldluecke2013variational}
Goldluecke, B., Wanner, S.: The variational structure of disparity and
  regularization of {4D} light fields. In: CVPR (2013)

\bibitem{gortler1996lumigraph}
Gortler, S.J., Grzeszczuk, R., Szeliski, R., Cohen, M.F.: The lumigraph. In:
  ACM SIGGRAPH (1996)

\bibitem{heber2014shape}
Heber, S., Pock, T.: Shape from light field meets robust {PCA}. In: ECCV (2014)

\bibitem{heber2013variational}
Heber, S., Ranftl, R., Pock, T.: Variational shape from light field. In: Energy
  Minimization Methods in Computer Vision and Pattern Recognition. Springer
  (2013)

\bibitem{isaksen2000dynamically}
Isaksen, A., McMillan, L., Gortler, S.J.: Dynamically reparameterized light
  fields. In: ACM SIGGRAPH. ACM (1996)

\bibitem{kim2013scene}
Kim, C., Zimmer, H., Pritch, Y., Sorkine-Hornung, A., Gross, M.H.: Scene
  reconstruction from high spatio-angular resolution light fields. ACM SIGGRAPH
   (2013)

\bibitem{levoy1996light}
Levoy, M., Hanrahan, P.: Light field rendering. In: ACM SIGGRAPH. ACM (1996)

\bibitem{li2014saliency}
Li, N., Ye, J., Ji, Y., Ling, H., Yu, J.: Saliency detection on light field.
  In: CVPR (2014)

\bibitem{liang2008programmable}
Liang, C.K., Lin, T.H., Wong, B.Y., Liu, C., Chen, H.H.: Programmable aperture
  photography: multiplexed light field acquisition. ACM SIGGRAPH  (2008)

\bibitem{lin2015depth}
Lin, H., Chen, C., Bing~Kang, S., Yu, J.: Depth recovery from light field using
  focal stack symmetry. In: ICCV (2015)

\bibitem{ng2006digital}
Ng, R.: Digital light field photography. Ph.D. thesis, stanford university
  (2006), [Ren Ng founded Lytro]

\bibitem{ng2005light}
Ng, R., Levoy, M., Br{\'e}dif, M., Duval, G., Horowitz, M., Hanrahan, P.: Light
  field photography with a hand-held plenoptic camera. Computer Science
  Technical Report CSTR  2(11) (2005)

\bibitem{park2011high}
Park, J., Kim, H., Tai, Y.W., Brown, M.S., Kweon, I.: High quality depth map
  upsampling for {3D-TOF} cameras. In: ICCV (2011)

\bibitem{perwassnext}
Perwass, C., Wietzke, L.: The next generation of photography.
  https://github.com/pcarbo/lbfgsb-matlab (2010), [Online; accessed
  15-April-2015; Perwass and Wietzke founded Raytrix]

\bibitem{perwass2012single}
Perwass, C., Wietzke, L.: Single lens {3D}-camera with extended depth-of-field.
  In: IS\&T/SPIE Electronic Imaging (2012)

\bibitem{sebe2002multi}
Sebe, I.O., Ramanathan, P., Girod, B.: Multi-view geometry estimation for light
  field compression. In: Annual Workshop on Vision, Modeling and Visualization:
  VMV (2002)

\bibitem{sun2010secrets}
Sun, D., Roth, S., Black, M.J.: Secrets of optical flow estimation and their
  principles. In: CVPR (2010)

\bibitem{tao2013depth}
Tao, M.W., Hadap, S., Malik, J., Ramamoorthi, R.: Depth from combining defocus
  and correspondence using light-field cameras. In: ICCV (2013)

\bibitem{tosic2014light}
Tosic, I., Berkner, K.: Light field scale-depth space transform for dense depth
  estimation. In: CVPR Workshops (2014)

\bibitem{vaish2004using}
Vaish, V., Wilburn, B., Joshi, N., Levoy, M.: Using plane+ parallax for
  calibrating dense camera arrays. In: CVPR (2004)

\bibitem{wang2015occlusion}
Wang, T.C., Efros, A.A., Ramamoorthi, R.: Occlusion-aware depth estimation
  using light-field cameras. In: ICCV (2015)

\bibitem{wanner2012globally}
Wanner, S., Goldluecke, B.: Globally consistent depth labeling of {4D} light
  fields. In: CVPR (2012)

\bibitem{wanner2012spatial}
Wanner, S., Goldluecke, B.: Spatial and angular variational super-resolution of
  {4D} light fields. In: ECCV (2012)

\bibitem{wanner2014variational}
Wanner, S., Goldluecke, B.: Variational light field analysis for disparity
  estimation and super-resolution. IEEE Transactions on Pattern Analysis and
  Machine Intelligence (PAMI)  36(3),  606--619 (2014)

\bibitem{wanner2013datasets}
Wanner, S., Meister, S., Goldluecke, B.: Datasets and benchmarks for densely
  sampled 4d light fields. In: Annual Workshop on Vision, Modeling and
  Visualization: VMV (2013)

\bibitem{wanner2013globally}
Wanner, S., Straehle, C., Goldluecke, B.: Globally consistent multi-label
  assignment on the ray space of {4D} light fields. In: CVPR (2013)

\bibitem{zhang2015light}
Zhang, Z., Liu, Y., Dai, Q.: Light field from micro-baseline image pair. In:
  CVPR (2015)

\bibitem{zhu1997algorithm}
Zhu, C., Byrd, R.H., Lu, P., Nocedal, J.: Algorithm 778: {L-BFGS-B}: Fortran
  subroutines for large-scale bound-constrained optimization. ACM TOMS  23(4),
  550--560 (1997)

\end{thebibliography}
\end{document}